\documentclass[12pt,onecolumn,journal,draftclsnofoot]{IEEEtran} 

\usepackage{algorithm}% 
\usepackage[noend]{algpseudocode}
\usepackage{graphicx}
\usepackage{graphicx}
\usepackage{graphics}
\usepackage{amssymb,mathrsfs}
\usepackage{amsmath,mathtools}

\usepackage{color}
\usepackage{xspace}
\usepackage{bbm}
\usepackage{comment}
\usepackage{algorithmicx}
\usepackage{subfig}
\usepackage{psfrag}
\usepackage{lipsum}
\usepackage{caption} 
\usepackage{tikz}
\usepackage[utf8]{inputenc}
\usepackage{amsmath}
\usepackage{bbm}
\usepackage{hyperref}
\definecolor{CranJ}{cmyk}{0,0.69,0.54,0.04} %cranberry jello
\definecolor{PinkJ}{cmyk}{0,0.71,0.43,0.12} %pink jeep
\definecolor{Cran}{cmyk}{0,0.73,0.41,0.29} %cranberry 
\definecolor{VRed}{cmyk}{0,0.75,0.25,0.2} %violetred
\definecolor{ORed}{cmyk}{0,0.75,0.75,0} %orangered4
\definecolor{CBlue}{cmyk}{1,0.25,0,0} %curacao	
 \newcommand{\reals}{{\mathbb{R}}}

\newcommand{\argmax}{\operatorname{argmax}}
\newcommand{\vect}[1]{\boldsymbol{\mathbf{#1}}}
\newcommand{\vectsf}[1]{\vect{\mathsf{#1}}}

\newcommand{\oprocendsymbol}{\hbox{$\bullet$}}
\newcommand{\oprocend}{\relax\ifmmode\else\unskip\hfill\fi\oprocendsymbol}

\def\BibTeX{{\rm B\kern-.05em{\sc i\kern-.025em b}\kern-.08emT\kern-.1667em\lower.7ex\hbox{E}\kern-.125emX}}
\begin{document}

\columnsep 6mm 

\title{Clustering of Time Series Data with Prior Geographical Information}

\author{Reza Asadi, and Amelia Regan%
\thanks{Reza Asadi and Amelia Regan are with department of Computer Science at University of California Irvine, USA, e-mail: \{rasadi, aregan\}@uci.edu}}%

%\markboth{IEEE Transactions on intelligent transport systems, ~Vol.~XX, No.~X, Nov~2020}%
%{Shell \MakeLowercase{\textit{et al.}}: Bare Demo of IEEEtran.cls for IEEE Journals}

\maketitle

\begin{abstract}
Time Series data are broadly studied in various domains of transportation systems. Traffic data are a challenging example of spatio-temporal data, as it is multi-variate time series with high correlations in spatial and temporal neighborhoods. Spatio-temporal clustering of traffic flow data find similar patterns in both spatial and temporal domain, where it provides better capability for analyzing a transportation network, and improving related machine learning models, such as traffic flow prediction and anomaly detection. In this paper, we propose a spatio-temporal clustering model, where it clusters time series data based on spatial and temporal contexts. We propose a variation of a Deep Embedded Clustering (DEC) model for finding spatio-temporal clusters. The proposed model Spatial-DEC (S-DEC) use prior geographical information in building latent feature representations. We also define evaluation metrics for spatio-temporal clusters. Not only do the obtained clusters have better temporal similarity when evaluated using DTW distance, but also the clusters better represents spatial connectivity and dis-connectivity. We use traffic flow data obtained by PeMS in our analysis. The results show that the proposed Spatial-DEC can find more desired spatio-temporal clusters.
\end{abstract}

\begin{IEEEkeywords}
Deep Learning, Time Series Clustering, Spatio-temporal data, Traffic Flow Data
\end{IEEEkeywords}
% \maketitle

\section{Introduction}

Spatio-temporal data arise in broad areas of engineering and environmental sciences. Data mining techniques have been used extensively for spatio-temporal analysis \cite{atluri2018spatio}. Geo-referenced time series are a subset of spatio-temporal data, where fixed locations over a geographical area observes some features for a time period in a synchronous way.  Traffic data is a complex example of Geo-referenced data, which is multi-variate
time series data, including the flow, speed
and occupancy of a large number of sensors, and in which there are correlations and similarities in spatial and temporal neighborhood. Spatio-temporal analysis of traffic data have a pivotal role in future research to improve the performance of transportation systems \cite{nagy2018survey}, such as reducing traffic congestion and air pollution \cite{chowdhury2017data}, understanding the behaviour of a transportation network \cite{rempe2016spatio}, predicting traffic speed and flow \cite{zang2018long}, \cite{asadi2020spatio}, and detecting non-recurrent congestion events \cite{anbaroglu2014spatio}. 

The volume and variety of spatio-temporal data has increased with the advent of new
sensing technologies, such as cameras, GPS and sensors \cite{toch2019analyzing}. Increases in the volume of traffic data requires the development of large-scale machine learning algorithms and big data analytics \cite{zhu2018big}, and data-driven approaches on traffic data \cite{wang2016soft}. Deep learning models have been recently successfully applied on spatial and temporal domains \cite{wang2019deep}. The models especially outperforms traditional machine learning and statistical methods on large-scale data. Several studies shows the success of deep learning solutions, such as traffic flow forecasting \cite{ma2020daily}, missing data imputation \cite{chen2019traffic} and spatio-temporal modelling of traffic flow data \cite{dixon2019deep}. Success of deep learning models in various domains along with the challenges of applying the deep learning models on spatio-temporal traffic data are the main motivations to further study the problem.

\subsection{Clustering of traffic data}

Spatio-temporal clustering of traffic data has been broadly studied with various goals. First, congestion detection and prediction can assist travelers and traffic management systems to improve the efficiency of existing systems. Second, detecting similarity in traffic patterns can help machine learning models to find similar regions in a transportation network.  This can improve missing data imputation and traffic forecasting models, or can identify anomalies in the data.

In \cite{cheng2018classifying}, they propose an improvement of fuzzy k-means clustering to classify traffic states into five groups ranging from mild to extreme traffic. Also, in \cite{celikoglu2014dynamic}, a clustering of traffic flow data is obtained based on congestion levels. They describe clusters in the temporal domain based on levels of congestion. While these works cluster traffic data based on traffic congestion, they did not consider spatial domains in their analysis. Moreover, in \cite{wei2020spatio}, they propose a method to better understand how traffic conditions are correlated in space-time. They cluster traffic data based on four congestion levels using an improved spatio-temporal Moran scatter-plot. These works cluster traffic data based on level of congestion. However, we consider clustering of traffic data based on similarities of patterns, which can be more generalizable to various machine learning problems, such as traffic flow prediction and anomaly detection. In \cite{anbaroglu2014spatio}, they define a measure, called the Link Journey Time, and they obtain spatio-temporal clusters of non-recurrent events. Each spatio-temporal cluster is a non-recurrent detected event, where it represents neighboring spatial and temporal features. Their model consider a similarity measure to obtain spatio-temporal data. However, their model is only designed to find non-recurrent events, and not generalized to find similar regions or temporal patterns. 

In \cite{shi2019detection}, they consider the problem of clustering of traffic flow data to obtain spatial and temporal similar patterns. They propose spatio-temporal clustering of traffic flow by considering topology of the network and similarity of time series data, where clusters are made by successive connections of neighbors. This work considers prior assumptions about the data and the topology of the network. A data-driven approach is expected to find spatio-temporal clusters without any prior assumptions, which would be more generalizable to different problems and scenarios \cite{kim2017data}. In \cite{cheng2007mining}, similarities of urban traffic flow are explored with a discrete wavelet transform. In \cite{chunchun2011traffic}, they proposed a fuzzy clustering method on traffic flow segments. Dynamic Time Warping (DTW), as a temporal similarity function, is used to identify locations with temporal similarities. They consider the problem of clustering of time series segments. In \cite{nguyen2019feature}, they represent spatio-temporal data as image-like representation. They propose a point-based and segment-based clustering of speed to represents classes of traffic congestion in spatial and temporal domain. A segment-based clustering is a similar approach to our model, where we find the clustering based on time series segments. However, they use a filter to obtain features from computer vision. On the other hand, we consider a temporal similarity distance to represent similarities in traffic flow data. Moreover, they evaluate clusters by visually assessing the model's output, where we use such a visualization method to represent interesting insight in the clusters of traffic flow data.

Similarity of traffic patterns not only detect traffic congestion, but also detect spatial and temporal heterogeneous neighborhoods. In \cite{tang2019short}, a k-means clustering is applied to find traffic flow variations based on spatio-temporal correlations. The clusters of similar locations are the input to a neural network which predicts traffic flow with higher performance. In \cite{ku2016clustering}, cluster of similar locations have been used with an autoencoder to impute missing values. In \cite{qiu2019traffic}, a clustering method finds road segments based on their features and missing data imputation is applied on incomplete speed data. In \cite{salamanis2017identifying}, a clustering model is used to identify anomalies in traffic flow data. We consider the problem  of discovering spatio-temporal similarities in traffic data. Clusters of traffic data, such as speed, flow and occupancy can represent levels of congestion. However, traffic flow data can be represents locations and time stamps with similar patterns with the goal of finding heterogeneous spatial and temporal domains.

\subsection{Spatio-temporal clustering with deep learning}

Since we consider clustering of traffic flow segments, here we review some of the recent research regarding time series clustering and in the rest, we describe the literature review for deep learning models for clustering problems. In \cite{aghabozorgi2015time}, they describe a broad range of time series clustering applications. The main components of time series clustering are studied including time series representations, similarity and distance measures, clustering prototypes and time series clustering. In \cite{soheily2016generalized}, they describe the challenges of k-means clustering with time warp measures. They propose weighted and kernel time warp measures for k-means clustering. Their method has a faster estimation of clusters. Further investigation of time series clustering is studied in \cite{paparrizos2017fast}. These works illustrate that novelty in time series representations and distance measures are the main approaches of improving temporal clustering.

There are a broad range of clustering models applied to spatio-temporal data, such as k-means \cite{huang2016time}, DBSCAN \cite{birant2007st}, agglomerative clustering \cite{yao2018stepwise}, and matrix factorization based clustering \cite{zhou2018visual}. However, increases in the size of datasets requires more scalable models such as deep learning models. When there is a huge dataset that includes data points with spatial and temporal properties, applying traditional clustering methods such as k-means on traffic data is computationally expensive and can have poor performance \cite{huang2016time}. More efficient heuristic methods for k-means clustering of traffic flow data have been studied \cite{tang2015hybrid}. Complex spatio-temporal patterns in traffic data necessitate further consideration of spatial and temporal information in the models. Deep learning models significantly improve performance of various machine learning problems, such as computer vision and natural language processing. Deep learning models have been broadly used for various large-scale spatio-temporal problems \cite{wang2019deep}, \cite{asadi2019convolution}. Moreover, deep learning models for clustering tasks are broadly studied in \cite{min2018survey}. Deep embedded clustering is primarily introduced in \cite{xie2016unsupervised}. Variations of the model have been studied in broad domains. Joint training of the model to preserve the latent feature space structure is proposed in \cite{guo2017improved}. In \cite{yang2017towards}, they analyze clustering-friendly latent representations, which jointly optimize dimension reduction using both a neural network and k-means clustering. While most of the research applies deep embedded clustering on images, there are few studies to show their performance on time series data. In \cite{tzirakis2019time}, they jointly cluster and train the model. They also segment time series data with agglomerative clustering. In \cite{8970987}, they propose a DEC with a cluster tree structure to dynamically obtain the number of clusters, while the original DEC has a fix number of clusters. However, these models do not consider any prior relation among the clusters. In \cite{ren2019semi}, they propose a variation of DEC, which considers the pairwise distance between data points. The model uses the prior distances as a measure to classify unlabeled data points. This work consider a relation among the clusters for unsupervised learning, and it is similar to our work, as we consider any prior relation among the clusters based on their geographical information.
In \cite{madiraju2018deep}, they evaluate various similarity metrics to obtain clusters with DEC. While these models consider temporal similarity in a DEC, there is a lack of development of a deep learning model where it not only finds clusters based on temporal similarity, but also prior spatial features would be considered.

\subsection{Contributions of the work}

The aforementioned works on clustering of time series data show the importance, advantages and applications of applying deep learning models to cluster time series and spatio-temporal traffic data. It also shows that recently there have been several deep learning models developed for clustering problems, where their goal is to modify latent feature space. There are various works that develop deep learning models for clustering of time series data. Clustering of time series data finds cluster of a transportation network based on the similarity of traffic flow data \cite{asadi2019spatio}. However, considering prior geographical information in designing clusters is a challenging problem.

In this paper, we focus on clustering of time series with prior geographical information. We propose a model where it modify latent feature representation based on geographical information. The model is a variation of DEC, where it finds spatio-temporal clusters by adding a new loss function to the model.

The contributions of the paper are as follows:

\begin{itemize}
    \item We formulate spatio-temporal clustering of traffic flow data as the clustering of time series segments.
    \item A spatial deep embedded clustering (Spatial-DEC) model is proposed which considers prior geographical information within the latent feature representation. To the best of our knowledge, this is the first work which considers prior geographical information to obtain spatial clusters with the DEC model.
    \item We illustrate the application clustering of traffic flow data in transportation systems.
    \item The spatio-temporal clusters obtained by deep learning models are evaluated on traffic flow data available in PeMS.
\end{itemize}

In section II, we describe the problem definition. In section III, we describe the technical background of the proposed model. In section IV, a deep learning model, Spatial-DEC, is proposed for spatio-temporal clustering. In Section V the models are evaluated on traffic flow data. Section VI describes the conclusions and future works.

%Here, we illustrate the main definitions and backgrounds of the work.

\section{Problem Definition}
Spatio-temporal data is represented with a three dimensional matrix $\vectsf{X} \in \reals^{\mathsf{s}, \mathsf{\overline{t}} , \mathsf{f}}$, where $\mathsf{s}$ is the number of sensors, $\mathsf{\overline{t}}$ is the number of time stamps and $\mathsf{f}$ is the number of traffic features, including flow, speed and occupancy. 
Each fixed location has its own multi-variate time series data $\vectsf{x}_{\mathsf{i}} \in \reals^{\mathsf{\overline{t}},\mathsf{f}}$. A sliding window method, given a time window $\mathsf{w}$, generates a sequence of data points. In other words, the function $\mathsf{slidingWindow}(.,.)$ receives input data $\vectsf{X}$ and time window size $\mathsf{w}$, and outputs data points $\vectsf{X}_\mathsf{d}$, which consists of all data points, time series segments, at time stamp $\mathsf{t}$ and location $\mathsf{i}$, represented with $\vectsf{x}_\mathsf{i}^\mathsf{t} \in \reals^{\mathsf{w},\mathsf{f}}$. Throughout the paper, we represent each data point with two indices $\mathsf{i}$ for location index and $\mathsf{t}$ for temporal index. A clustering method assigns a data point $\vectsf{x}_\mathsf{i}^\mathsf{t}$ into a cluster $\vectsf{c}_\mathsf{j}$, where $\mathsf{j} \in \{1, \dots, \mathsf{c}\}$, and $\mathsf{c}$ is the given number of clusters. While in alternative approaches, one can consider the problem of clustering of the whole time series $\vectsf{x}_{\mathsf{i}} \in \reals^{\overline{\mathsf{t}},\mathsf{f}}$, e.g. clustering of trajectory data \cite{sabarish2018clustering}, or sub sequences of spatio-temporal data $\vectsf{x}^{\mathsf{t}} \in \reals^{\mathsf{s},\mathsf{w},\mathsf{f}}$. A clustering model finds similar data points based on a distance function, such as euclidean distance. Here, we define a \textit{temporal cluster}, when its members have high temporal similarity, which can be obtained with a DTW distance function. It is desirable to have a more dense and compact temporal cluster. We define a \textit{spatial cluster}, which includes location indices where their data points have high temporal similarity. 

\begin{figure}[t]\hspace{-0.0in}%[htbp]
  \centering   
  {
    \includegraphics[height=1.8in]{ 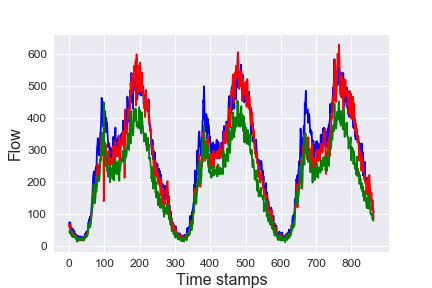}
  }\\ %goes to next line
  \hspace{-0.0in}
  \caption{An example of traffic flow data for three road segments, represented with three colors. The time stamps are every 5-min and the figure represents traffic flow data for three days.}\vspace{-0.1in}\label{fig::spatialC}
\end{figure}

\begin{figure}[t]
\centering
\begin{tikzpicture}[auto,thick,scale=0.8, every
    node/.style={scale=0.8}]
    \tikzstyle{mynode}=%
    [%
    minimum size=12pt,%
    inner sep=0pt,%
    outer sep=0pt,%
    draw,
    %ball color=gray!10,%example text.fg,%
    shape=circle%
    ]
    
    \node[inner sep=0pt] (l1) at  (4.0,2.3)
    {\text{input data points} };
    \draw [draw=black] (8.5,-2.1) rectangle (-1,2.0);
    
    \draw [draw=black] (2.0,1.9) rectangle (-0.3,0.1);
    \node[inner sep=0pt] (33) at  (0.9,1)
    {\includegraphics[width=0.12\textwidth]{ 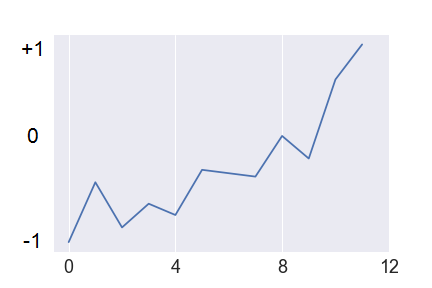}};
    \node[inner sep=0pt] (l1) at  (-0.6,1.0)
    {\text{\large $\mathsf{x}_1^{\mathsf{t}}$} };

    \draw [draw=black] (2.0,-0.1) rectangle (-0.3,-1.9);
    \node[inner sep=0pt] (33) at  (0.9,-1.0)
    {\includegraphics[width=0.12\textwidth]{ 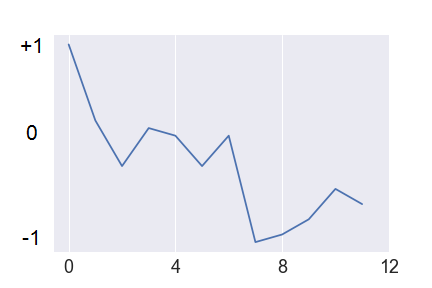}};
    \node[inner sep=0pt] (l1) at  (-0.6,-1.0)
    {$\mathsf{x}_1^{\mathsf{t}+1}$};

    \draw [draw=black] (5.2,1.9) rectangle (2.8,0.1);
    \node[inner sep=0pt] (33) at  (4,1)
    {\includegraphics[width=0.12\textwidth]{ 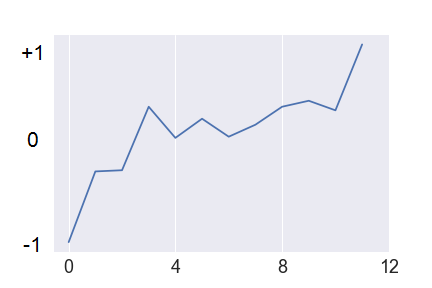}};
    \node[inner sep=0pt] (l1) at  (2.5,1.0)
    {\text{\large $\mathsf{x}_2^{\mathsf{t}}$} };

    \draw [draw=black] (5.2,-0.1) rectangle (2.8,-1.9);
    \node[inner sep=0pt] (33) at  (4,-1)
    {\includegraphics[width=0.12\textwidth]{ 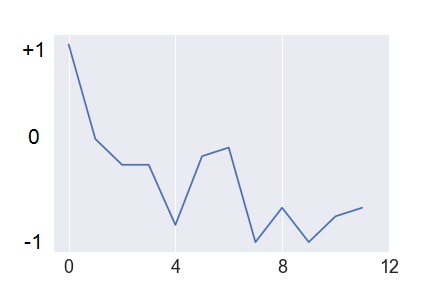}};
    \node[inner sep=0pt] (l1) at  (2.5,-1.0)
    {$\mathsf{x}_2^{\mathsf{t}+1}$};

    \draw [draw=black] (8.4,1.9) rectangle (6.0,0.1);
    \node[inner sep=0pt] (33) at  (7.2,1)
    {\includegraphics[width=0.12\textwidth]{ 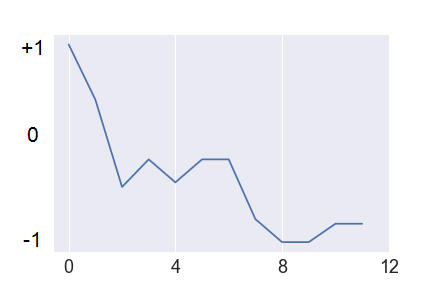}};
    \node[inner sep=0pt] (l1) at  (5.7,1.0)
    {\text{\large $\mathsf{x}_3^{\mathsf{t}}$} };
    
    \draw [draw=black] (8.4,-0.1) rectangle (6.0,-1.9);
    \node[inner sep=0pt] (33) at  (7.2,-1)
    {\includegraphics[width=0.12\textwidth]{ 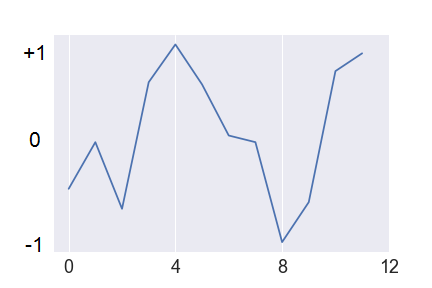}};
    \node[inner sep=0pt] (l1) at  (5.7,-1.0)
    {$\mathsf{x}_3^{\mathsf{t}+1}$};

    \node[inner sep=0pt] (l1) at  (1.0,-5.2)
    {\text{Cluster 1} };
    \node[inner sep=0pt] (l1) at  (4,-5.2)
    {\text{Cluster 2} };
    \node[inner sep=0pt] (l1) at  (7,-5.2)
    {\text{Cluster 3} };
    
    \draw [draw=black] (1.9,-11.0) rectangle (-1.4,-5.5);
    \draw [draw=black] (5.3,-9.2) rectangle (2.0,-5.5);
    \draw [draw=black] (8.7,-7.4) rectangle (5.4,-5.5);
    
    \draw [draw=black] (1.8,-7.3) rectangle (-0.6,-5.7);
    \node[inner sep=0pt] (33) at  (0.6,-6.5)
    {\includegraphics[width=0.12\textwidth]{ 444.png}};
    \node[inner sep=0pt] (l1) at  (-1,-6.5)
    {\text{\large $\mathsf{x}_1^{\mathsf{t}+1}$} };
    
    \draw [draw=black] (1.8,-9.1) rectangle (-0.6,-7.5);
    \node[inner sep=0pt] (33) at  (0.6,-8.3)
    {\includegraphics[width=0.12\textwidth]{ 555.png}};
    \node[inner sep=0pt] (l1) at  (-1,-8.3)
    {\text{\large $\mathsf{x}_2^{\mathsf{t}+1}$} };
    
    \draw [draw=black] (1.8,-10.9) rectangle (-0.6,-9.3);
    \node[inner sep=0pt] (33) at  (0.6,-10.1)
    {\includegraphics[width=0.12\textwidth]{ 666.png}};
    \node[inner sep=0pt] (l1) at  (-1,-10.1)
    {\text{\large $\mathsf{x}_3^{\mathsf{t}}$} };
    
    \draw [draw=black] (5.2,-7.3) rectangle (2.8,-5.7);
    \node[inner sep=0pt] (33) at  (4,-6.5)
    {\includegraphics[width=0.12\textwidth]{ 111.png}};
    \node[inner sep=0pt] (l1) at  (2.4,-6.5)
    {\text{\large $\mathsf{x}_1^{\mathsf{t}}$} };
    \draw [draw=black] (5.2,-9.1) rectangle (2.8,-7.5);
    \node[inner sep=0pt] (33) at  (4,-8.3)
    {\includegraphics[width=0.12\textwidth]{ 222.png}};
    \node[inner sep=0pt] (l1) at  (2.4,-8.3)
    {\text{\large $\mathsf{x}_2^{\mathsf{t}}$} };
    
    \draw [draw=black] (8.6,-7.3) rectangle (6.2,-5.7);
    \node[inner sep=0pt] (33) at  (7.4,-6.5)
    {\includegraphics[width=0.12\textwidth]{ 333.png}};
    \node[inner sep=0pt] (l1) at  (5.8,-6.5)
    {\text{\large $\mathsf{x}_3^{\mathsf{t+1}}$} };

    \draw [draw=black] (6.5,-4.0) rectangle (1.5,-3);
    
    \node[inner sep=0pt] (l1) at  (4,-3.5)
    {\text{Spatio-temporal clustering} };

    \node[inner sep=0pt] (111) at  (4.0,-2.1){};
    \node[inner sep=0pt] (222) at  (4.0,-3.0){};
    \draw[thick, ->] (111) --  (222);
    
    \node[inner sep=0pt] (111) at  (4.0,-4.1){};
    \node[inner sep=0pt] (222) at  (4.0,-5.1){};
    \draw[thick, ->] (111) --  (222);
    
    \node[inner sep=0pt] (111) at  (4.0,-4.1){};
    \node[inner sep=0pt] (222) at  (2.0,-5.1){};
    \draw[thick, ->] (111) --  (222);
    
    \node[inner sep=0pt] (111) at  (4.0,-4.1){};
    \node[inner sep=0pt] (222) at  (6.0,-5.1){};
    \draw[thick, ->] (111) --  (222);

    \end{tikzpicture} 
    \caption{An example of spatio-temporal clustering of traffic flow data}
    \label{fig::spatiotempoal-clustering}
\end{figure}

Traffic flow data is a spatio-temporal data. In Fig. \ref{fig::spatialC}, we represent an example of traffic flow for three sensors and three days. Finding temporal similarity among road networks is challenging with point-wise clustering. To prevent from the fluctuations in the clusters, we consider segment-wise clustering. In Fig \ref{fig::spatiotempoal-clustering}, we describe a schematic representation of the input and output of the spatio-temporal clustering. The input data points $\vectsf{x}_\mathsf{i}^\mathsf{t}$ are time series segments for location $\mathsf{i}$, three road segments, and time stamp $\mathsf{t}$, two time stamps. The selected data points are from PeMS traffic flow data, but the time stamps and road segments are arbitrary and are presented with the purpose of clarification of the problem definition. Each data point is a time series of length 12 in the figure. For 5-min time stamps, each data point represents one hour of traffic flow data for one road segment. The horizontal axis is time stamps, and vertical axis is traffic flow, normalized in range $[-1, +1]$. The three output clusters represents similar data points. The clusters represent similar patterns over different days and hours. They also represents the locations that are similar on a transportation networks.

%A spatial cluster represents road segments where they have high similarity of traffic data. The first approach is to cluster whole time series and find fixed spatial clusters. However, road segments can have dynamic similarities over temporal domain. The second approach is to obtain spatial clusters based on temporal similarities. In this paper, we obtain spatial clusters based on similarities of traffic data segments. In this approach, dynamic spatial clusters can be obtained over different time periods. Desirable spatial clusters find locations connected with each other, and are more compact. It should be evident that a disconnected cluster of road segments is meaningless for spatial similarities. %Hence, in this paper, we propose a clustering method to obtain spatio-temporal clusters, where our goal is to find temporal clusters with high temporal similarities, while the clusters are dense and compact. Also, our second goal is to find spatial clusters, which they have high connectivity, and low dis-connectivity.

%Spatio-temporal clustering of traffic data is a challenging problem, since the clustering method not only should consider repeated temporal patterns as a result of seasonality and trends, but also consider geographical information and relation of road networks. The data also includes noise, where it can fluctuate the clusters. Hence, we require to have a segment-wise clustering model where it can consider temporal similarity of traffic flow segments.

This spatio-temporal clustering problem is challenging. First, the clustering method should consider both temporal and spatial similarities. Second, given a large number of time stamps, e.g. six months, $\overline{\mathsf{t}}$ and a large number of road segments of a city $\mathsf{s}$, a sliding window method generates $\overline{\mathsf{t}}\times\mathsf{s}$ of data points, which can be a very large number of data points. While k-means clustering methods have been proposed for time series segments, their performance drops when they faced with large number of data points, and it has expensive computational time. Moreover, a k-means clustering method has some limitations to be modified and consider both spatial and temporal similarities. Hence, in this work, we propose a deep learning model, Spatial-DEC, to solve spatio-temporal clustering problem.

\section{Technical Background}

\subsection{Autoencoders}
An autoencoder is primarily proposed in \cite{vincent2010stacked}. It consists of an encoder $\phi(\vectsf{x}) = \sigma(\mathsf{drop}(\vectsf{x})\vectsf{w}^1+\vectsf{b}^1)$ and a decoder $\theta(\vectsf{z}) = \sigma(\mathsf{drop}(\vectsf{z})\vectsf{w}^2+\vectsf{b}^2)$, where the activation function and the dropout function are represented with $\sigma(.)$ and $\mathsf{drop}(.)$, respectively.
An encoder is the first neural network component, which reduces the dimension of input data to a latent feature space $\vectsf{z} \in \reals^\mathsf{d}$, where $\mathsf{d} < \mathsf{n}$. The second neural network component is a decoder, which reconstructs the input data from its latent representation.

In a deep autoencdoer, the encoder and decoder consist of several layers. The encoder and decoder are a symmetric and multi-layered neural network. The loss function, e.g. mean square error, reduces the difference of input data and its reconstruction. In other words, the input and target data are both $\vectsf{X}_\mathsf{d}$. For the given spatio-temporal data, the reconstruction loss function is as follows,

\begin{align}
    \sum_{\mathsf{t}=1}^{\overline{\mathsf{t}}}\sum_{\mathsf{i}=1}^\mathsf{s} \frac{1}{2} ||\vectsf{y}_\mathsf{i}^\mathsf{t} - \theta(\phi(\vectsf{x}_\mathsf{i}^\mathsf{t}))||^2
\end{align}

where $\overline{\mathsf{t}}$ is the number of time stamps, and $\mathsf{s}$ is the number of sensors or locations. Also, $\vectsf{y}_\mathsf{i}^\mathsf{t}$ is the target of an autoencoder, which is the same as input data $\vectsf{x}_\mathsf{i}^\mathsf{t}$. Minimization of this objective function results in learning the latent feature representation of input data. We consider weight of $\alpha_2$ for reconstruction loss throughout the paper, and in our representation of autoencoders the weight is $\alpha_2=1$.

\subsection{Deep Embedded Clustering}
A Deep embedded clustering neural network is introduced in \cite{xie2016unsupervised}. The encoder transforms $\vectsf{x}$ into latent feature space $\vectsf{z}$. The clustering layer is connected to latent feature layer. The weights of clustering layer are initialized with cluster centers obtained by k-means clustering. Cluster center $\mathsf{j}$ is represented with $\mu_\mathsf{j} \in \reals^{\mathsf{d}}$. Given $\mathsf{k}$ as the number of clusters, and $\mathsf{d}$ as latent feature size, the clustering layer is represented with a dense layer $\reals^{\mathsf{d}} \rightarrow \reals^{\mathsf{k}}$. In other words, it converts latent features into a vector of size $\mathsf{k}$, which $\mathsf{j}$-th element represents the probability that the data point is assigned to the cluster $\mathsf{j}$.

Given initial cluster centers $\{\mu_1, \dots, \mu_\mathsf{k}\}$, obtained by k-means clustering, and latent features $\vectsf{z}$, a student's t-distribution measures the similarity between cluster centers $\mu_\mathsf{j}$ and data points $\vectsf{x}_\mathsf{i}$ as follows,
%\vspace{-0.05in}
\begin{align}
    \mathsf{q}_{\mathsf{i}\mathsf{j}} = \frac{(1+||\mathsf{z}_\mathsf{i} - \mu_\mathsf{j}||^2)^{-1}}{\sum_\mathsf{k}(1+||\mathsf{z}_\mathsf{i} - \mu_\mathsf{k}||^2)^{-1}}
\end{align}

where the degree of freedom of the student's t-distribution is one. The probability of assigning a data point $\vectsf{x}^\mathsf{t}_\mathsf{i}$ to a cluster with center point $\mu_\mathsf{j}$ is represented with $\mathsf{q}^\mathsf{t}_{\mathsf{i}\mathsf{j}}$. The assigned cluster is $\argmax_{\mathsf{j}} \vectsf{q}^\mathsf{t}_{\mathsf{i}\mathsf{j}}$. The clustering algorithm iteratively adjusts clusters by learning from high confidence assignments. To learn from high confidence assignments, an auxiliary target distribution $\mathsf{p}_{\mathsf{i}\mathsf{j}}$ is as follows,
%\vspace{-0.2in}
\begin{align}
    \mathsf{p}^\mathsf{t}_{\mathsf{i}\mathsf{j}} = \frac{(\mathsf{q}^\mathsf{t}_{\mathsf{i}\mathsf{j}})^2/\mathsf{f}_{\mathsf{j}}} {\sum_\mathsf{k} (\mathsf{q}^\mathsf{t}_{\mathsf{i}\mathsf{j}})^2/\mathsf{f}_\mathsf{k} },
\end{align}

where $\mathsf{f}_{\mathsf{j}}$ is the number of elements in cluster $\mathsf{j}$.
KL-divergence loss between $\mathsf{q}^\mathsf{t}_{\mathsf{i}\mathsf{j}}$ and $\mathsf{p}^\mathsf{t}_{\mathsf{i}\mathsf{j}}$ learns the high confidence soft cluster assignment,

\begin{align}
    \mathsf{KL}(\mathsf{P}||\mathsf{Q}) = \sum_\mathsf{t} \sum_\mathsf{i} \sum_\mathsf{j} \mathsf{p}^\mathsf{t}_{\mathsf{i\mathsf{j}} } \log \frac{\mathsf{p}^\mathsf{t}_{\mathsf{i}\mathsf{j}}}{\mathsf{q}^\mathsf{t}_{\mathsf{i}\mathsf{j}}}
\end{align}

In \cite{guo2017improved}, they train the DEC with joint learning of clustering loss and reconstruction loss. In joint training, the loss function of the neural network on spatio-temporal data is as follows,

\begin{align}
    \sum_{\mathsf{t}=1}^{\overline{\mathsf{t}}}\sum_{\mathsf{i}=1}^\mathsf{s} (\sum_{\mathsf{j}=1 }^\mathsf{c}\alpha_1 \mathsf{p}^\mathsf{t}_{\mathsf{i\mathsf{j}} } \log \frac{\mathsf{p}^{\mathsf{t}}_{\mathsf{i}\mathsf{j}} }{\mathsf{q}^\mathsf{t}_{\mathsf{i}\mathsf{j}}} + \frac{\alpha_2}{2} ||\vectsf{y}_\mathsf{i}^\mathsf{t} - \theta(\phi(\vectsf{x}_\mathsf{i}^\mathsf{t})||^2) 
    % + \sum_{\mathsf{k}=\mathsf{i}\times \mathsf{s}}^{(\mathsf{i}+1)\times \mathsf{s}}~~(\frac{\vectsf{\overline{\lambda}}_{\mathsf{i}\mathsf{k}}}{2}||\phi(\vectsf{x}_\mathsf{i}^\mathsf{t}) - \vectsf{z}^\mathsf{t}_\mathsf{k} ||^2))
\end{align}

where $\mathsf{c}$ is the given number of clusters, $\mathsf{\alpha_2}$ is the weight of mean square error term and $\mathsf{\alpha_1}$ is the weight of clustering loss term. Minimization of the loss function in equation (5) results in learning the latent feature representation and the output clusters. The model receives input data $\vectsf{X}_\mathsf{d}$ and target data are $\vectsf{p}$ for clustering layer and $\vectsf{X}_\mathsf{d}$ for decoder's output. For DEC, the value of $\alpha_1$ and $\alpha_2$ represents the importance of each of loss functions. Higher value of $\alpha_1$ reduces loss function for clustering, while higher value of $\alpha_2$ better keep the structure of autoencoder's latent features \cite{guo2017improved}.

\section{Spatial Deep Embedded Clustering}

Here, we describe the proposed method for spatio-temporal clustering of traffic data. Algorithm 1 is the procedure of finding spatio-temporal clusters on traffic flow data. 

A deep embedded clustering (DEC) receives data points, $\vectsf{x}_{\mathsf{i}}^\mathsf{t}$ for all locations $\mathsf{i}$ and time stamps $\mathsf{t}$, represented with $\vectsf{X}_\mathsf{d}$. The encoder transforms each data point to its latent feature representation $\vectsf{z}_{\mathsf{i}}^\mathsf{t}$. Given the number of clusters $\mathsf{k}$, a k-means clustering on latent feature representations finds mean of the clusters for latent features. The mean of a cluster $\vectsf{\mu}_\mathsf{j}$ is obtained by k-means clustering and is stored into the clustering layer.

\begin{figure}[t]\hspace{-0.0in}%[htbp]
  \centering   
  
    \includegraphics[height=2.2in]{ 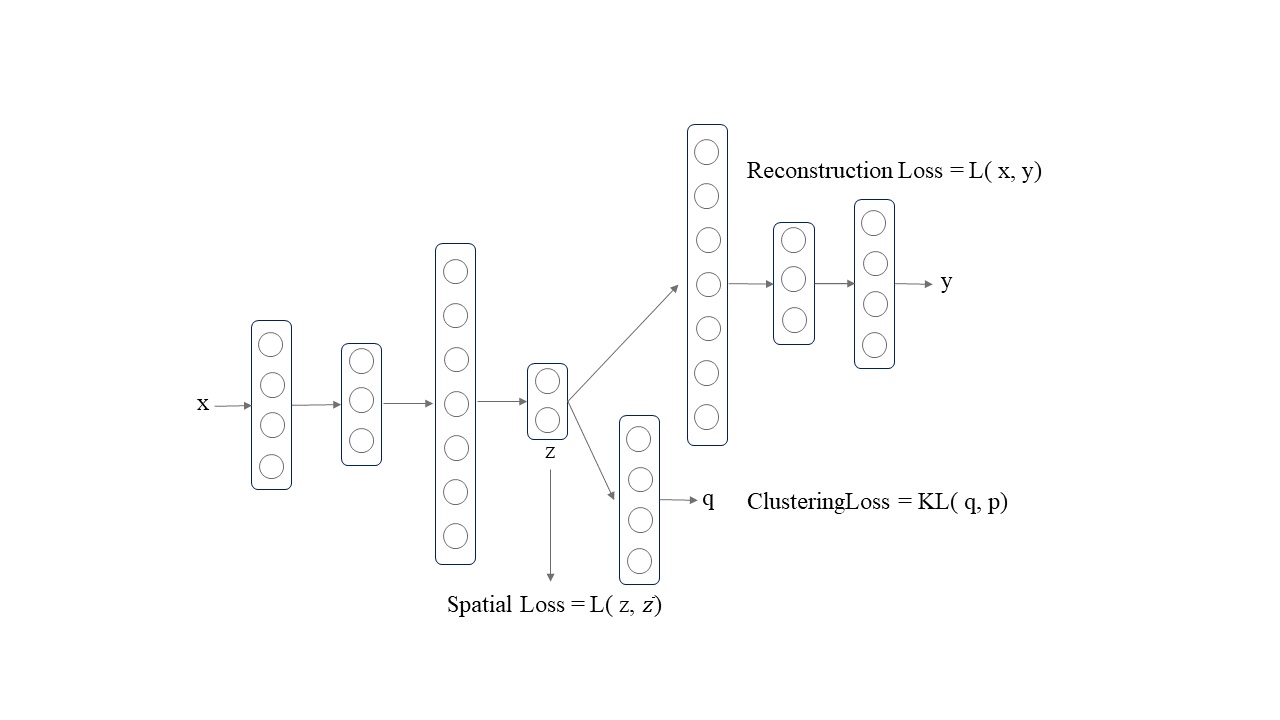}
  \caption{The architecture of the Spatial-DEC. The input data $\vectsf{x}$ has target values $[\vectsf{\overline{z}}, \vectsf{p}, \vectsf{y}]$.}
  \vspace{-0.1in}\label{fig::architecture}
\end{figure}

The data points with high temporal similarity are close to each other in the latent feature space, examined in Fig. \ref{fig::latent_tsne11}. Hence, each cluster includes data points with high temporal similarity. However, not only the clusters should represents data points with high temporal similarity, but also they should consider data points of spatial neighborhood. If a cluster represent data points of locations, far from each other or distributed in a geographical area, it is not our desired cluster. Hence, our objective is to obtain clusters with both temporal similarity and spatial closeness. In the rest of this section, we describe our modification to deep embedded clustering and introduce Spatial-DEC, its architecture is presented in Fig. \ref{fig::architecture}. The objective is to modify the DEC's loss function, so if $\vectsf{x}_\mathsf{i}$ and $\vectsf{x}_\mathsf{j}$ are close (or far from) each other, then their latent representations $\vectsf{z}_\mathsf{i}$ and $\vectsf{z}_\mathsf{j}$ are also close (or far from) each other.

First, we make the latent feature representations conditional to the prior geographical location. The proposed model needs the location indices as the input data, because it maps the data points to latent feature space based on both their time series values and location indices. For a given $\mathsf{s}$ sensors, we generate a one-hot encoding of the locations $\vectsf{g} \in \reals^{\mathsf{s}}$, where the location $\mathsf{i}$ has the input vector $\vectsf{g}$, in which $\vectsf{g}_\mathsf{i} = 1$ and the rest of the values are zeros. This computation of one-hot encoding is obtained  $\mathsf{OnehotEncoding}()$ in the algorithm 1. The S-DEC receives $[\vectsf{x}_\mathsf{i}^\mathsf{t}, \vectsf{g}_\mathsf{i}]$ as the input data. The encoder $\phi([\vectsf{x}_\mathsf{i}^\mathsf{t},\vectsf{g}_\mathsf{i}])$ outputs latent feature representations $\vectsf{z}_\mathsf{i}^\mathsf{t}$. Given the time series segment and its location the encoder outputs low dimensional representation of the data.

Next, we add a loss function into the DEC, and propose Spatial-DEC (S-DEC), where its latent features are constructed given prior geographical information. We add spatial loss term, $\mathsf{sl}(\phi([\vectsf{x}, \vectsf{g}]) , \vectsf{\overline{z}})$, to the latent feature layer. The encoder's input are $\vectsf{x}$ and $\vectsf{g}$. The encoder's output is $\phi([\vectsf{x}, \vectsf{g}])$. The encoder's output in the last training steps is stored in $\vectsf{\overline{z}}$. In other words, the model uses $\vectsf{\overline{z}}$ as the target value for the latent feature layer. In Algorithm (1), it is shown that once the DEC obtains $\vectsf{p}$ as the target value for the clustering layer, we also obtains $\vectsf{\overline{z}}$ as the target value for latent feature layer $\vectsf{z}$.

To implement the new loss function effectively, first we change the size of the input data. The goal is to have input data $\vectsf{x}_\mathsf{i}^\mathsf{t}$ and target data $\vectsf{\overline{z}}_\mathsf{k}^\mathsf{t}$ for each pair of locations $\mathsf{i}$ and $\mathsf{k}$. We repeat each data point, a row in $\vectsf{X}_\mathsf{d}$, $\mathsf{s}$ times. The new training data are stored in $\vectsf{\overline{X}}_\mathsf{d} \in \reals^{\mathsf{s}\times\mathsf{s}, \mathsf{w}}$. Moreover, we reshape the latent features of the last training step, $\vectsf{\overline{z}}$. Each block of $\mathsf{s}$ rows of the $\vectsf{z}$ are repeated $\vectsf{s}$ times. The reshaped latent features are represented with $\vectsf{\overline{z}} \in \reals^{\mathsf{s}\times\mathsf{s}, \mathsf{w}}$. The Spatial-DEC has $\vectsf{\overline{X}}_\mathsf{d}$ as an input data point, and $\vectsf{\overline{z}}$ as the target for latent feature layer. After changing the size of input data and target data, for any given $[\vectsf{x}_\mathsf{i}^\mathsf{t},\vectsf{g}_\mathsf{i}]$ and its encoder's output $\phi(\vectsf{x}_\mathsf{i}^\mathsf{t})$, there are target data $\vectsf{\overline{z}}_\mathsf{k}^\mathsf{t}$ for all $\mathsf{k}$. This modification allows the model to control the distance of latent features at location $\mathsf{i}$ with all locations $\mathsf{k}$. The loss function optimizes the distance of $\vectsf{z}_\mathsf{i}^\mathsf{t}$ to all previously obtained latent features at the same time stamp, represented with $\vectsf{\overline{z}}_\mathsf{k}^\mathsf{t}$ for all sensors $\mathsf{k}$. The loss function should increase (decrease) the distance of $\vectsf{z}_\mathsf{i}^\mathsf{t}$ and $\vectsf{\overline{z}}_\mathsf{k}^\mathsf{t}$, if location $\mathsf{i}$ and $\mathsf{k}$ are far from (close) each other. In the rest, we define the weight matrix $\vectsf{\overline{\lambda}}$ which controls the distance of latent features.

We define a weight value for the loss function, represented with $\vectsf{\overline{\lambda}}_{\mathsf{i}\mathsf{k}}$, where it is weight that represent the distance of two locations $\mathsf{i}$ and $\mathsf{k}$. A transportation network can be represented with a graph, and $\vectsf{\overline{\lambda}}$ is the adjacency matrix that represents distances of locations. Here we assume that all locations are on a line without loss of generality. We define $\vectsf{\lambda} \in \reals^{\mathsf{s}, \mathsf{s}}$ as the weight matrix, which represents the spatial distance of locations, where $\vectsf{\lambda}_{\mathsf{i}\mathsf{k}} \in [-1, 1]$ represents the distance of two locations $\mathsf{i}$ and $\mathsf{k}$. If the value is close to +1, then two locations $\mathsf{i}$ and $\mathsf{k}$ are close each other, and if the value is close to -1, then two locations are far from each other. Given $\vectsf{\lambda}_\mathsf{i} \in [-1,1]^{\mathsf{1},\mathsf{s}}$ as the i-th row of $\vectsf{\lambda}$, we define $\vectsf{\lambda}^\mathsf{d}_\mathsf{i} \in [-1,+1]^{\mathsf{s},\mathsf{s}}$ as the diagonal matrix of the elements of $\vectsf{\lambda}_\mathsf{i}$, that is all element except diagonal elements are zero. We obtain $\overline{\vectsf{\lambda}} \in \reals^{\mathsf{s} \times \mathsf{s}, \mathsf{s}}$ by aggregating $\vectsf{\lambda}^\mathsf{i}_\mathsf{d}$ for all locations $\mathsf{i}$ on the first dimension. The calculation is represented with $\mathsf{LossWeight}(.)$ in the algorithm 1. The spatial loss term is as follows,

\begin{equation} \sum_{\mathsf{t}=1}^{\overline{\mathsf{t}}}\sum_{\mathsf{i}=1}^{\mathsf{s}}\sum_{\mathsf{k}=\mathsf{i}\times \mathsf{s}}^{(\mathsf{i}+1)\times \mathsf{s}}~~(\frac{\vectsf{\overline{\lambda}}_{\mathsf{i}\mathsf{k}}}{2}||\phi([\vectsf{x}_\mathsf{i}^\mathsf{t}, \vectsf{g}_\mathsf{i}]) - \vectsf{\overline{z}}^\mathsf{t}_\mathsf{k} ||^2)
\end{equation}

In a back-propagation method, the gradients of the Equation (6) along with the gradient of autoencoder's loss function are propagated in the neural network. We only describe the back-propagation for Equation (6). We refer the reader to \cite{xie2016unsupervised} for further theoretical analysis of gradient propagation for DEC. Given $\mathsf{sl}$ as the spatial loss function, the gradient of Equation (6) is as follows,

\begin{equation} \frac{\partial \mathsf{sl}}{\partial \vectsf{z}_\mathsf{i}^\mathsf{t}}=\frac{\partial (\frac{\vectsf{\overline{\lambda}}_{\mathsf{i}\mathsf{k}}}{2}||\vectsf{z}_\mathsf{i}^\mathsf{t} - \vectsf{\overline{z}}^\mathsf{t}_\mathsf{k} ||^2) }{\partial \vectsf{z}_\mathsf{i}^\mathsf{t}} = \vectsf{\overline{\lambda}}_{\mathsf{i}\mathsf{k}}( \vectsf{z}_\mathsf{i}^\mathsf{t} - \vectsf{\overline{z}}^\mathsf{t}_\mathsf{k})
\end{equation}

\begin{algorithm}[t!]
  \caption{Spatio-temporal clustering with S-DEC }\label{alg:algo_dtw}
  \begin{algorithmic}[1]
    \Procedure{Clustering}{$~\vectsf{X} = \{\vectsf{x}_1, \dots, \vectsf{x}_\mathsf{s} \}$}
    \State $\vectsf{X}_\mathsf{d} \gets \mathsf{slidingWindow}(\vectsf{X}, \mathsf{w})$
    \State $\vectsf{g} \gets \mathsf{OnehotEncoding}()$
    \State $\vectsf{\overline{\lambda}} = \mathsf{LossWeight}()$
    \State $\mathsf{modelAutoencoder}.\mathsf{trainAutoencoder}(\vectsf{x} = [\vectsf{X}_\mathsf{d},\vectsf{g}], \vectsf{y} = \vectsf{X}_\mathsf{d})$
    \State $\vectsf{z} \gets \mathsf{modelAutoencoder}.\mathsf{predictLatent}([\vectsf{X}_\mathsf{d}, \vectsf{g}])$
    \State $\vectsf{\mu} \gets \mathsf{initialKmeans}(\vectsf{z})$
    \State $\mathsf{modelSDEC} \gets \mathsf{initializeDEC}(\vectsf{\mu}, \mathsf{modelAutoencoder})
    $
    \State $\vectsf{\overline{X}}_\mathsf{d} \gets \mathsf{resizex}(\vectsf{X}_\mathsf{d})$
    %\State $\mathsf{modelEmb} = \mathsf{trainEmbC}(\mathsf{modelPre} ,[\vectsf{x}^1, \dots, \vectsf{x}^\mathsf{\overline{t}}], [\mu^1, \dots, \mu^\mathsf{c}])$
    %\State \Comment{End of training step.}
    
    % \State $\vectsf{\overline{z}},\vectsf{q},\vectsf{y} \gets \mathsf{modelSDEC}.\mathsf{predictClusters}([\vectsf{x},\vectsf{g}])$
    % \State $\vectsf{p} \gets \mathsf{targetDist}(\vectsf{q})$
    
    % \For{ $[\vectsf{xb},\vectsf{zb},\vectsf{pb}] \in [\vectsf{x},\vectsf{\overline{z}}, \vectsf{p}]$ 
    % \State $\mathsf{modelSDEC}.\mathsf{train}(\vectsf{x} = [\vectsf{xb}, \vectsf{g}], \vectsf{y} = [\vectsf{zb}, \vectsf{pb}, \vectsf{xb}])$
    % \EndFor}
    
    \For{ $\mathsf{epoch} \in [1,~\dots,~ \mathsf{maxEpoch}]$ 
    \State $\vectsf{z},\vectsf{q},\vectsf{y} \gets \mathsf{modelSDEC}.\mathsf{predictClusters}([\vectsf{X}_\mathsf{d},\vectsf{g}])$
    \State $\vectsf{\overline{z}} \gets \mathsf{resizez}(\vectsf{z})$
    \State $\vectsf{p} \gets \mathsf{targetDist}(\vectsf{q})$
        \For{ $[\vectsf{xb},\vectsf{zb},\vectsf{pb}] \in [\vectsf{X}_\mathsf{d},\vectsf{\overline{z}}, \vectsf{p}]$ 
        \State $\mathsf{modelSDEC}.\mathsf{train}(\vectsf{x} = [\vectsf{xb}, \vectsf{g}], \vectsf{y} = [\vectsf{zb}, \vectsf{pb}, \vectsf{xb}])$
        \EndFor}
    \EndFor}
    
    %\State $\vectsf{Q} = \mathsf{modelEmb}.\mathsf{predictClusters}([\vectsf{x}^1, \dots, \vectsf{x}^\mathsf{\overline{t}}])$
    % \For{ $\mathsf{i} \gets 1$ to $\mathsf{s}$ 
    %     \For{ $\mathsf{j} \gets \mathsf{i}$ to $\mathsf{s}$
    %     \State$\vectsf{C}[\mathsf{i},\mathsf{j}] = \sum_{\mathsf{t} \in \vectsf{T}} (\vectsf{Q}^\mathsf{t}_{\mathsf{i}} == \vectsf{Q}^\mathsf{t}_{\mathsf{j}})$
    %     \EndFor}
    % \EndFor}
    \State \textbf{return} $\vectsf{modelSDEC}$ \Comment{\small Return the trained spatial-DEC model for spatio-temporal clusteirng of traffic data}
    \EndProcedure 
  \end{algorithmic}
\end{algorithm}

The model finds the gradient with respect to the encoder's output $\vectsf{z}_\mathsf{i}^\mathsf{t}$. In stochastic gradient descent algorithm, the gradient is propagated to update the weights of the neural network. The loss function is similar to \cite{ren2019semi}. They uses pairwise distances among clusters and apply the model on clustering of images for unsupervised learning.

Here we describe the reason that the value of $\vectsf{\lambda}_{\mathsf{i}\mathsf{k}}$ directly affects the structure of latent features and the clustering model. The encoder's output for a given input data point $\vectsf{x}_\mathsf{i}^\mathsf{t}$ is $\vectsf{z}_\mathsf{i}^\mathsf{t}$. The value of encoder's output for last training step and same time stamp $\mathsf{t}$ is stored in $\vectsf{\overline{z}}^\mathsf{t}$. Given the data point at location $\mathsf{i}$, the model considers a target value $\vectsf{z}_\mathsf{k}^\mathsf{t}$ for all $\mathsf{k}$. Since the neural network minimizes the loss function, the value of $\vectsf{\lambda}_{\mathsf{i}\mathsf{k}}$ controls the distance $\vectsf{z}_\mathsf{i}^\mathsf{t}$ given the last estimation of $\vectsf{\overline{z}}_\mathsf{k}^\mathsf{t}$ for all $\mathsf{k}$. If $\vectsf{\lambda}_{\mathsf{i}\mathsf{k}}$ has a positive value, then the loss value is positive for the distance of $\vectsf{z}_\mathsf{i}^\mathsf{t}$ and $\vectsf{\overline{z}}_\mathsf{k}^\mathsf{t}$. Training a neural network with such loss value reduces the distance of latent features of $\vectsf{z}_\mathsf{i}^\mathsf{t}$ and $\vectsf{\overline{z}}_\mathsf{k}^\mathsf{t}$. On the other hand, a negative value for $\vectsf{\lambda}_{\mathsf{i}\mathsf{k}}$ increases the distance of latent features of $\vectsf{z}_\mathsf{i}^\mathsf{t}$ and $\vectsf{\overline{z}}_\mathsf{k}^\mathsf{t}$. In section 5.2, we validate the effect of new loss function on a sample data. The weight for spatial loss function, represented in Equation (6), is $\alpha_0$. In the experimental results, $\alpha_0$, $\alpha_1$ and $\alpha_2$ are the weights of spatial loss, Equation 6, clustering loss, Equation 4, and reconstruction loss, Equation 1, respectively.

In algorithm (1), we describe procedure of finding spatio-temporal clusters. Lines 2-9 are the preprocessing of a DEC model, which includes pretraining an autoencoder, k-means initialization, and building a DEC model. In line 5, the pre-training of autoencoder is with loss weight values of $\alpha_0 = 1.0$, $\alpha_1 = 0.0$ and $\alpha_2 = 1.0$. In line 11, the model finds value of latent features for the last pretraining step, and it follows by obtaining $\vectsf{\overline{z}}$ and $\vectsf{p}$ as target values. Line 13 is the function in Equation (3), introduced in DEC. Line 14 generates batch size for input and output data. To clarify our notation, we illustrate the input arguments of a neural network with $\vectsf{x}$, and output arguments with $\vectsf{y}$, represented in $\mathsf{model}.\mathsf{train}(\vectsf{x}, \vectsf{y})$.

Lastly, we analyze the computational time of the model. In traffic flow data, we have a large number of data points, $\mathsf{n} = \mathsf{s}\times\mathsf{\bar{t}}$, where it is the multiplication of the number of location and total time stamps. A k-means clustering method finds the clusters in order of $O(\mathsf{n}^2)$ steps. In each step, the method requires to calculates DTW distance, where it is $O(\mathsf{w}^2)$
for time series segments of $\mathsf{w}$. The DEC model maps time series segments to a lower dimension $\mathsf{l}$. In our experiments, we consider $\mathsf{w} = 12$ and $\mathsf{l}=4$. A DEC model finds k-means initialization by a sub-sample of data points, and train the model in $O(\mathsf{E}\mathsf{n})$ steps, where $\mathsf{E}$ is the number of epochs. With mini-batch gradient descent we expect to train the model in less than 100 epochs based on our experiments. In each step, a DEC model requires to apply back-propagation, where its computational time depends on size of the neural network.

\section{Experimental Results}
Here we illustrate the results for clustering of traffic flow data. The deep learning model is implemented with Keras. We use a fully-connected autoneocder with 7 layers. All of the layers have Relu activation function and dropout rate of 0.2. The number of hidden units are $(8, 8, 128, 4, 128, 8, 8)$ in seven fully-connected layers. The batch size of 288, one day with 5-min time stamp, and Adam optimizer are selected. 

We compare the performance of three models, k-means on latent features of an autoencoder, DEC and Spatial-DEC. We also have three loss terms, spatial loss, clustering loss, and reconstruction loss, with corresponding weights $\alpha_0$, $\alpha_1$ and $\alpha_2$, respectively. For k-means with autoencoder, we have following weights, $\alpha_0 = 0$, $\alpha_1 = 0$ and $\alpha_2 = 1$. For DEC, we have $\alpha_0 = 0$, $\alpha_1 = 0.2$ and $\alpha_2 = 1$. For Spatial-DEC, we have $\alpha_0 = 0.1$, $\alpha_1 = 0.2$ and $\alpha_2 = 1$.

% The implementation of Spatial-Deep Embedded Clustering (S-DEC) on time series data is available here: \url{https://github.com/rezaa89/spatio\textunderscore temporal\textunderscore clustering}

\subsection{Traffic data}
\begin{figure}[ht]\hspace{-0.0in}%[htbp]
  \centering   
  
    \includegraphics[height=1.8in]{ 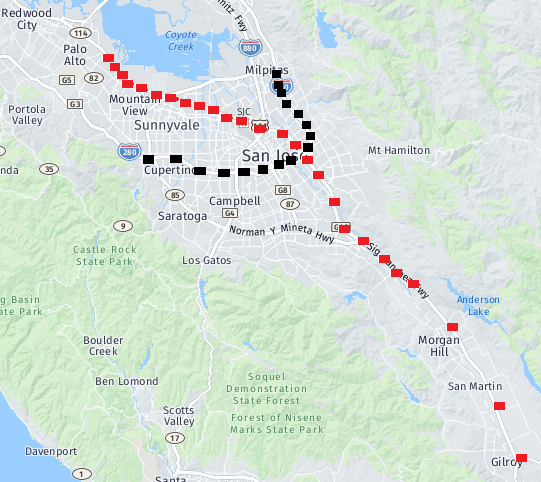}
  \caption{The main-line loop detector sensors on highways in Bay Area, California. The black boxes are the sensors on I-680-S and I-280-S, and red boxes are US101-S.}
  \vspace{-0.1in}\label{fig::map11}
\end{figure}

Traffic flow data are obtained from the PeMS~\cite{californiapems}. Traffic flow data is gathered from main-line loop detector sensors every 30 seconds and aggregated to every 5 minutes. The data are for US-101 South and, I-280 South and I-680 South highways, in the Bay Area of California, which includes 26 and 16 mainline sensors, respectively, illustrated in Fig. \ref{fig::map11}. We represent the average of values on these selected data. We select the data for the first five months of 2016. The models are trained on the first three months, and evaluated on the next two months. In a preprocessing step, we re-scale the data into the range of $[0,1]$, and subtract each time window of size $\mathsf{w}$ from its mean value. A time window of size 12, one hour, is selected. In the model we assume that sensors are on one line in a highway, and the average of results for these two highways is presented.

\subsection{Validation of the spatial loss function}
In Spatial-DEC, we use the spatial loss function introduced in Equation 6. The loss function decreases (increases) the latent feature representations, if two locations are close (far from) each other. Here, we examine the correctness of the model by visualizing the latent feature representations. We consider the first six successive sensors on the highway US101-S.

\begin{figure}[ht]\hspace{-0.0in}%[htbp]
  \centering   
  \subfloat[]{
    \includegraphics[height=1.1in]{ 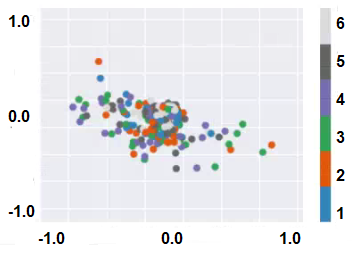}\label{fig::sample11}
  } %goes to next line
  \subfloat[]{
    \includegraphics[height=1.1in]{ 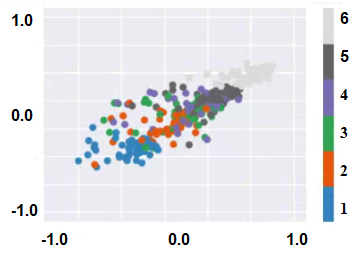}\label{fig::sample21}
  }\\ 
  \subfloat[]{
    \includegraphics[height=1.1in]{ 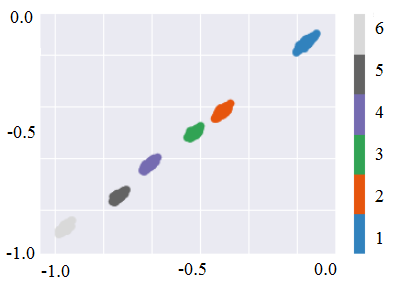}\label{fig::sample31}
  }\\ 
  \hspace{-0.0in}
  \caption{The scatter plot of latent feature representations for the sample data. Fig (a) represents the latent features of an autoencoder without spatial loss. Fig (b) represents the latent features with spatial loss of weight $\alpha_0 = 1$, and $\alpha_0 = 10$ in Fig (c).}\vspace{-0.1in}\label{fig::latent-sample}
\end{figure}

We obtain the spatial weights, which represents distances of location on a line. We need any arbitrary function to obtain weights in following ways. If two locations are close (far), their weights should be close to +1 (-1). We assume that there are six locations on one line. We use a distance function of $\lambda_{\mathsf{i}\mathsf{j}} = (1-2\times |\frac{\mathsf{i}-\mathsf{j}}{\mathsf{s}+1}|)$ for $\mathsf{i} \neq \mathsf{j}$, and zero for $\lambda_{\mathsf{i}\mathsf{i}}$. Throughout the experiment, we notice that it better stabilize the clusters. As an example, the spatial weights for location one are $\lambda_1 = [0, 0.71, 0.42, 0.14. -0.14, -0.42]$, and for location two are $\lambda_2 = [0.71, 0, 0.71, 0.42, 0.14, -0.14]$.% If we set the elements at $\lambda_{\mathsf{i}\mathsf{i}}$ to one, data points for each location becomes more compact and dense.

We train the Spatial-DEC on a sample data. Fig. \ref{fig::latent-sample} represents the latent features in two dimensional with different values of $\alpha_0$. We set $\alpha_0$ to 0, 1.0, 10.0, in Fig. 5.a, 5.b and 5.c respectively. With $\alpha_0 = 0.0$, the model is an autoencoder without spatial loss. The data points are scattered in latent feature space without any prior geographical information. With a higher value of $\alpha_0 = 1.0$, we train the Spatial-DEC model. The results is in Fig. 5.b, where the order of locations from 1 to 6 is preserved in the latent feature space. For a spatial loss weight of $\alpha_0 = 10$, the latent features are completely separable. The data points are mapped into latent feature with their corresponding order of distances, from top left to bottom right. In the rest of implementations, we use a the weight $\alpha_0 = 0.1$. it can be more comparable with DEC and k-means based on temporal similarity and it also finds latent feature with spatial closeness. We also notice that an early stopping of the deep learning results can prevent from completely separable data points like Fig 5.c, and results in Fig 5.b.

\begin{table}[t]\centering
\begin{tabular}{ |p{2.0cm}||p{2.0cm}|p{1.5cm}|p{1.5cm}|  }
 \hline
 \multicolumn{4}{|c|}{Clustering evaluation (percentage results)} \\
 \hline
Models & Sum Square Error to Mean of Clusters  & Connectivity & dis-connectivity\\
 \hline
  \hline
 kmeans   &   0.27  & 0.19 & 0.21 \\
  \hline
kmeans-DTW   &  0.28   &  0.17 & 0.19 \\
 \hline
DEC   &  \textbf{0.22}   &  0.19 & 0.18  \\
 \hline
Spatial-DEC   &  0.23   &  \textbf{0.45} & 0.11  \\
 \hline
 \hline
%   \hline
% KNN-FC-ED   &     & \\
%  \hline
% KNN-LSTM  &     & \\
%  \hline
% KNN-BiLSTM  &     & \\
%  \hline
% KNN-CNN-BiLSTM   &     & \\
% \hline
 \end{tabular}
 \caption{The comparisons of clustering models}\label{tab::results}
\end{table} 
%The analysis shows that the spatial loss represented in Equation (6) can consider prior geographical information in the latent feature representation. Without loss of generality, we assume that the locations are on one line. Since the clustering is applied on latent features, the clusters would be based on spatial closeness. The Spatial-DEC model uses the spatial loss function to find spatio-temporal clusters.

% \subsection{Performance metrics}
% Following performance metrics are defined to evaluate the clusters. First, we want to have clusters of size more than one, so there would be a similarity among neighboring time series. Second, we want a high penalty for sensors which are far from each other and they are in same cluster. In other words, the clusters would be small and not distributed in a geographical area.

% \begin{align}
%     \mathsf{Dis} &= \sum_\mathsf{i} \sum_\mathsf{j} (\mathsf{i}-\mathsf{j})\times\vectsf{S}_{\mathsf{i}\mathsf{j}}\\
%     \mathsf{L} &= \sum_\mathsf{i}\sum_\mathsf{j} \mathsf{w}_{\mathsf{i}\mathsf{j}} || \vectsf{x}^\mathsf{i} - \vectsf{\mu}^\mathsf{j} ||^2 
% \end{align}

% The first measure $\mathsf{Dis}$ represents the dissimilarity as a function of distance of two sensors. The second measure is clustering loss $\mathsf{L}$, as the distance of data points from the centers of the clusters, where $\mathsf{w}_{\mathsf{i}\mathsf{j}}$ is one if element is in the cluster, otherwise it is zero. The output matrix represents $\vectsf{S}_{\mathsf{i}\mathsf{j}}$ as the number of time stamps, two sensors $\mathsf{i}$ and $\mathsf{j}$ are in same cluster. 

\subsection{Analysis of temporal clusters}
\begin{figure}[ht]\hspace{-0.0in}%[htbp]
  \centering   
  {
    \includegraphics[height=1.7in]{ 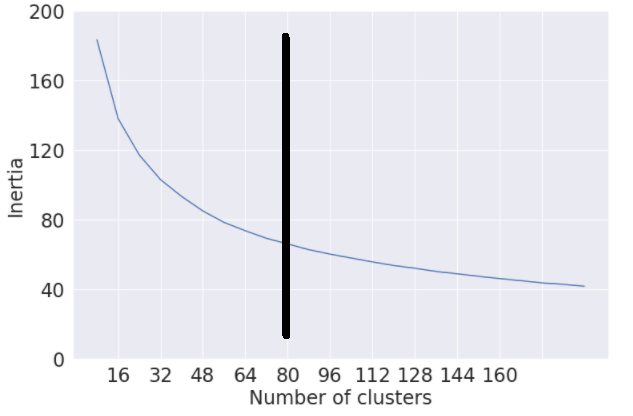}
  }\\ %goes to next line
  \hspace{-0.0in}
  \caption{The Elbow method finds the best value for the number of clusters represented with a black vertical line.}\vspace{-0.1in}\label{fig::inertia}
\end{figure}

After pretraining the autoencoders, the first step in training Spatial-DEC is to initialize the clusters with k-means clustering. To obtain an appropriate number of clusters, we use an Elbow method, i.e. the optimum value can be obtained,
when the reduction in inertia, as the
sum of squares of data points, becomes linear, represented in Fig. \ref{fig::inertia}, where we find 80 as the best number of clusters.

\begin{figure}[ht]\hspace{-0.0in}%[htbp]
  \centering   
  {
    \includegraphics[height=1.7in]{ 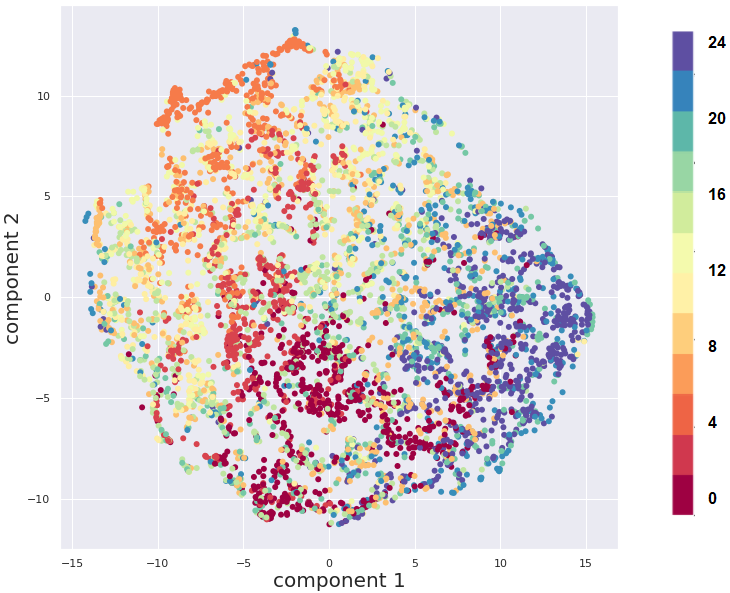}
  }
  \hspace{-0.0in}
  \caption{TSNE representation of autoencoder's latent features.}\vspace{-0.1in}\label{fig::latent_tsne11}
\end{figure}

To show that latent features are directly related to temporal features, the latent feature representation of one sensor's traffic states is shown for five weekdays in Fig. \ref{fig::latent_tsne11}. A t-distributed stochastic neighbor embedding (TSNE) \cite{maaten2008visualizing} method is used for representing latent features in two dimension, with parameters of 40, 300, 500 for preplexity, number of iterations and learning rate, respectively. The color of each data point represents the hours of a day. One day is grouped into 10 colors, for every 2 hours. This visualization of latent features show that the data points are distinguishable based on their time stamps and latent feature preserve temporal properties of data.

% \begin{figure}[h]\hspace{-0.0in}%[htbp]
%   \centering   
 
%   {
%     \includegraphics[height=1.9in]{Clustering/ corr_DTW_L.png}\label{fig::oneday}
%   }\\ %goes to next line
%   \hspace{-0.0in}
%   \caption{The plot for the relation of DTW distance and latent feature space.}\label{fig::corr}\vspace{-0.1in}
% \end{figure}

To represent temporal similarity in traffic flow data, Dynamic Time Warping have been broadly used \cite{lv2020temporal}. In our problem, warping window size can be obtained in the range of 1 to 12, the size of time series segments, where its smaller value reduces the computational time. Comparing the value of Rand Index based on warping window is a method to obtain the best value of warping window \cite{dau2018optimizing}. We notice that there is not a significant change in the clusters, obtained by kmeans clustering with DTW distance function, when we reduce warping window size from 12 to 6. Hence, we selected six as the best warping window. To show that latent feature space preserve the temporal distances, for any given data points, the euclidean distance
of latent features is calculated. Also the DTW among their
time series is calculated. The correlation
between latent features and dynamic time warping is obtained. For latent feature size from 1 to 10, the correlation changes from 0.9 to 0.98. The maximum  correlation
between dynamic time warping and euclidean distance of
latent features is 0.98 with latent feature 4. Hence, we selected a latent feature of size 4 in our analysis. We also conclude that the latent feature space preserve temporal similarity of data points.

In the rest, we analyze the clusters obtained by k-means, DEC and Spatial-DEC. Unlike supervised learning, in unsupervised learning there is not a clear approach to evaluate clusters. Hence, we describe properties that we expect to see in the clusters, and based on them define evaluation metrics. We expect to have clusters that they are compact and include data points with high temporal similarities. A feature-based clustering of time series data can improve the time series forecasting performance \cite{bandara2020forecasting}. Our clustering models are feature-based, where the clustering method is applied on the latent feature representation of data points. Hence, we define a temporal similarity measure as follows,

$$
\mathsf{s}^\mathsf{d}_\mathsf{j} = \frac{1}{|\vectsf{C}_\mathsf{j}|}\sum_{\vectsf{x}_\mathsf{i}^\mathsf{t} \in \vectsf{C}_\mathsf{j}} \mathsf{dtw}(\vectsf{x}_\mathsf{i}^\mathsf{t}, \vectsf{\bar{\mu}}_\mathsf{j})
$$

where $\vectsf{C}_\mathsf{j}$ is the set of members of cluster $\mathsf{j}$, $\bar{\mu}_\mathsf{j}$ the mean of the cluster and $\vectsf{x}_\mathsf{i}^\mathsf{t}$ is the element $\mathsf{i}$ of the cluster. We consider the element which its latent feature is the closest to the mean of the cluster as $\bar{\mu}$ or the mediods of the cluster. 

% Also, the inertia of k-means clusters is obtained as follows,

% $$
% \mathsf{s}^\mathsf{l}_\mathsf{j} = \frac{1}{|\vectsf{C}_\mathsf{j}|}\sum_{\vectsf{x}_\mathsf{i}^\mathsf{t} \in \vectsf{C}_\mathsf{j}} (\vectsf{z}_\mathsf{i}^\mathsf{t}- \vectsf{\mu}_\mathsf{j})^2
% $$

% where $\vectsf{\mu}_\mathsf{j}$ is the mean of the cluster $\mathsf{j}$ in latent feature space.

% Second, we expect to have a balanced size for the clusters \cite{malinen2014balanced}. We define a measure called as variation of size as follows,

% $$
% \mathsf{v} = \sum_{j=1}^{\mathsf{k}}
% \sum_{\vectsf{x}_\mathsf{i}^\mathsf{t} \in \vectsf{C}_\mathsf{j}} (|\vectsf{C}_\mathsf{j}| - \bar{\mathsf{s}})^2
% $$

% where $\mathsf{k}$ is the total number of clusters, and $\bar{\mathsf{s}}$ is the average size of clusters.

\begin{figure}[h]\hspace{-0.0in}%[htbp]
  \centering   
  {
    \includegraphics[height=1.9in]{ 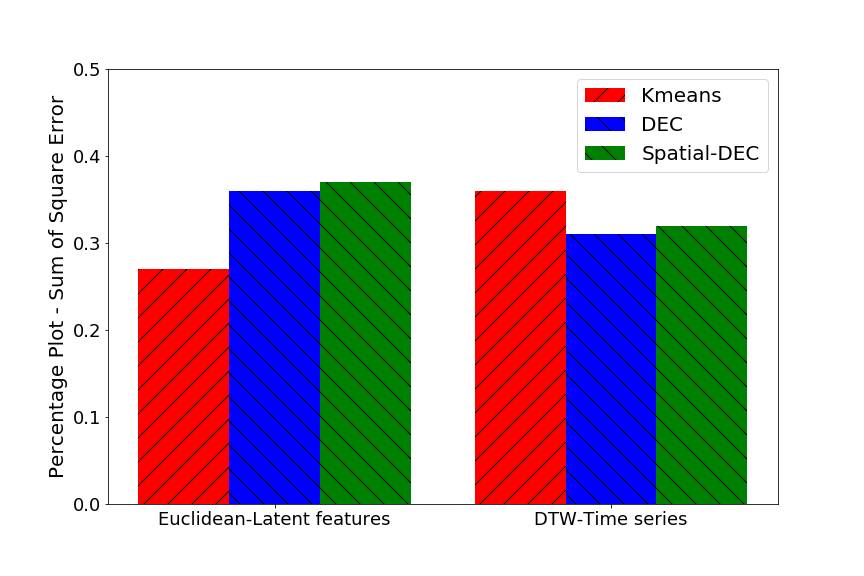}
  }
  \hspace{-0.0in}
  \caption{Comparison of implemented clustering models based on sum of square distance of clusters.}\vspace{-0.1in}\label{fig::comparison1}
\end{figure}

In Fig. \ref{fig::comparison1}, we compare the compactness of the clusters using $\sum_\mathsf{j} \mathsf{s}_\mathsf{j}^\mathsf{d}$ for DTW-time series, which shows the compactness of clusters based on temporal similarity. A more compact cluster better represents data points with temporal similarity. We compare the implemented DEC and Spatial-DEC, where both have similar temporal similarity. It is important to show that while Spatial-DEC finds more connected clusters, it does not significantly reduces temporal similarity in the clusters. The other model is k-means, applied on latent feature of time series. %We should mention that the best results can be obtained by k-means with DTW on original time series with SSE-DTW of 2400. However, it's computational complexity is significantly higher than the rest of the methods, and there is not trivial solutions to find clusters with high spatial connectivity. %Lastly, we compare the value of $\mathsf{v}$, as a measure to find balanced datasets.

\subsection{Analysis of spatial clusters}

Here, we evaluate spatial clusters obtained by Spatial-DEC. We define a spatial cluster as the set of locations which all have a similar assigned temporal cluster. In other words, for a given time stamp $\mathsf{t}$, all locations of one spatial cluster have equal assigned cluster $\mathsf{c}$. In traffic flow data analysis, a spatial cluster represents road segments with similar traffic flow patterns for a given time stamp. We define a connected spatial cluster $\vectsf{\delta}_\mathsf{i}^{\mathsf{t}}$, which includes location indices which have similar assigned cluster in a given time stamp $\mathsf{t}$ and they all are neighbors. We define \textit{spatial connectivity} as a evaluation metric for the analysis of spatial clusters. A spatial connectivity shows the total size of connected spatial clusters. If the size of connected spatial cluster of location $\mathsf{i}$ is one, then it means that the assigned cluster of location $\mathsf{i}$ is not equal to the assigned cluster of its neighbors. Such a clustering output is not desirable, as it cannot show similarity of locations. A desired spatial cluster should have high temporal similarity and spatial connectivity. We also mention that a high spatial connectivity can reduce temporal similarity, because the cluster includes larger road segments, which can have lower temporal similarity. A spatial connectivity is defined as follows,

%\begin{equation}
\begin{align}
\mathsf{s}_{\mathsf{c}} &= \sum_{\mathsf{t}=1}^\mathsf{\overline{t}} \sum_{\mathsf{i}=1}^\mathsf{s}  |\delta_\mathsf{i}^\mathsf{t}|
\end{align}
%\end{equation}

where $|\mathsf{\delta}|$ is the size of connected spatial cluster. 

On the other hand, if a spatial cluster includes location indices, dis-connected in a geographical area, then the cluster is not desirable. We define a evaluation metric \textit{spatial dis-connectivity} as follows. For each location $\mathsf{i}$ and time stamp $\mathsf{t}$, we define $\mathsf{\overline{\delta}}_\mathsf{i}^\mathsf{t}$, as the set of location indices which have an equal temporal cluster to $\mathsf{c}_\mathsf{i}^\mathsf{t}$, but they are not in $\mathsf{\delta}_\mathsf{i}^\mathsf{t}$. The spatial dis-connectivity is obtained as follows.

%\begin{equation}
\begin{align}
\mathsf{s}_{\mathsf{d}} &= \sum_{\mathsf{t}=1}^\mathsf{\overline{t}} \sum_{\mathsf{i}=1}^\mathsf{s}  |\overline{\delta}_\mathsf{i}^\mathsf{t}|
\end{align}
%\end{equation}

\begin{figure}[h]\hspace{-0.0in}%[htbp]
  \centering   
  {
    \includegraphics[height=1.7in]{ 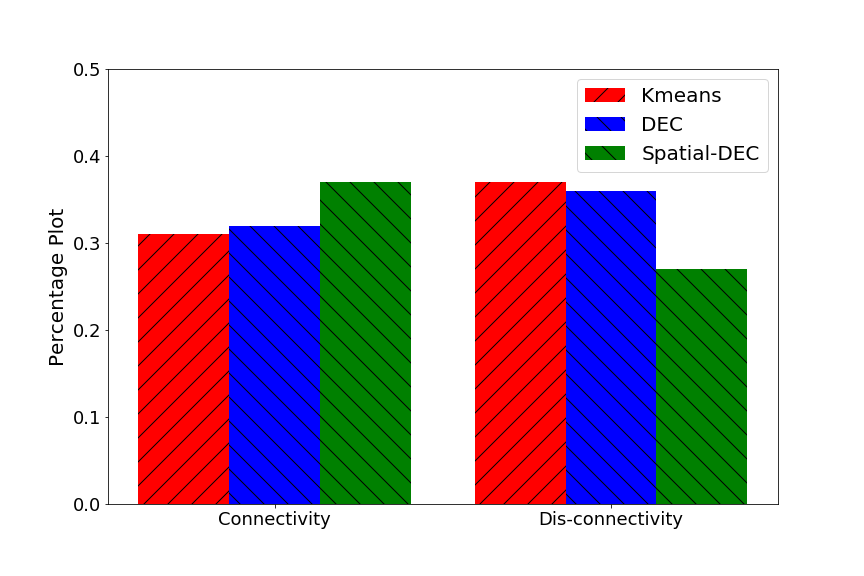}
  }
  \hspace{-0.0in}
  \caption{Comparison of implemented clustering models based spatial connectivity and dis-connectivity.}\vspace{-0.1in}\label{fig::comparison2}
\end{figure}

Fig \ref{fig::comparison2} represents the spatial connectivity and dis-connectivity of the obtained clusters. Higher value of connectivity shows that data points of closer locations are assigned into same temporal clusters, which is a more desired cluster. On the other hand, lower value of dis-connectivity represents that the clusters are not dis-connected in a geographical area. The figure shows that Spatial-DEC can significantly increase connectivity, and decreases the dis-connectivity of clusters.

Overall, Fig. 9 shows that the clusters of Spatial-DEC are more compact than k-means in terms of temporal similarity. In other words, all the data points of one cluster have higher temporal similarity. 

We define spatial metric as $\mathsf{s}_\mathsf{m} = \mathsf{s}_\mathsf{c} - \mathsf{s}_\mathsf{d}$. Here, we represent that the mean of $\mathsf{s}^\mathsf{t}_\mathsf{m}$ for all $\mathsf{t}$ obtained by Spatial-DEC is significantly higher than DEC. The null hypothesis $\mathsf{H}_0$ is that the mean of Spatial-DEC and DEC is equal. The alternate hypothesis $\mathsf{H}_\mathsf{a}$ is that their mean is not equal. Since, $\mathsf{s}_\mathsf{m}$ can be represented with a normal distribution with positive mean for both DEC and Spatial-DEC. We apply a t-test on $\mathsf{s}_\mathsf{m}$ obtained by DEC and Spatial-DEC. The p-value is 0.0012, where we can reject the null-hypothesis with significant level of $\alpha = 0.05$. It shows that the increase in connectivity of spatial clusters is statistically significant.

\subsection{Analysis of traffic flow clusters}

\begin{figure}[h]\hspace{-0.0in}%[htbp]
  \centering   
  {
    \includegraphics[height=1.9in]{ 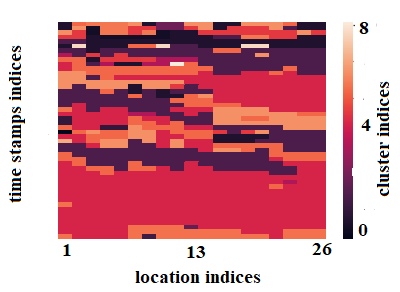}
  }
  \hspace{-0.0in}
  \caption{Spatio-temporal clusters obtained by Spatial-DEC}\vspace{-0.1in}\label{fig::regional_clusters_temporal}
\end{figure}

Here we visualize and further analyze clusters of traffic flow data. Spatio-temporal clusters shows that how road segments are similar
over time periods, represented in Fig \ref{fig::regional_clusters_temporal}. The figure shows time stamps for one day on y-axis and location indices on x-axis. Each color represents the assigned cluster. To better visualize clusters and represents their similarities, we only consider 8 clusters. The areas with same colors have temporal similarity. The figure shows how 26 locations are similar over time periods.

% Aggregating spatio-temporal clusters represents spatial clusters. In Fig. \ref{fig::regional_clusters}, a heatmap of the number similarity is represented, where it shows the number of times two locations have similar cluster in one time stamp. Higher values shows higher similarity in the heatmap. The result is for one week, and using a threshold the areas with highest values are selected as similar regions. There are totally nine clusters, nine rectangles, where it shows how sensors on I-101-S are similar.

% \begin{figure}[h]\hspace{-0.0in}%[htbp]
%   \centering   
%   {
%     \includegraphics[height=1.9in]{ similarity_hm.png}
%   }
%   \hspace{-0.0in}
%   \caption{Spatial clusters of I-101-S }\vspace{-0.1in}\label{fig::regional_clusters}
% \end{figure} 

\begin{figure}[h]\hspace{-0.0in}%[htbp]
  \centering   
  {
    \includegraphics[height=1.7in]{ 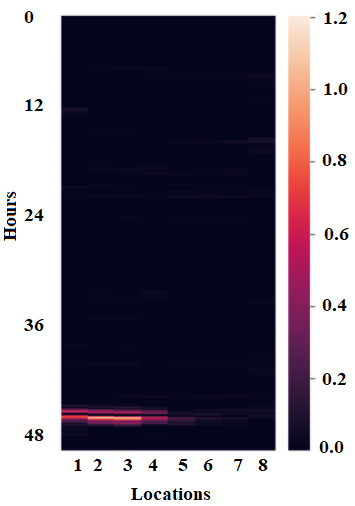}
  }
  \hspace{-0.0in}
  \caption{The heatmap of distance of data points to the center of their assigned clusters. The light colors represents data points far from center of clusters  and potential anomalies in data.}\vspace{-0.1in}\label{fig::mean_distance}
\end{figure}

\begin{figure}[h]\hspace{-0.0in}%[htbp]
  \centering   
  {
    \includegraphics[height=1.5in]{ 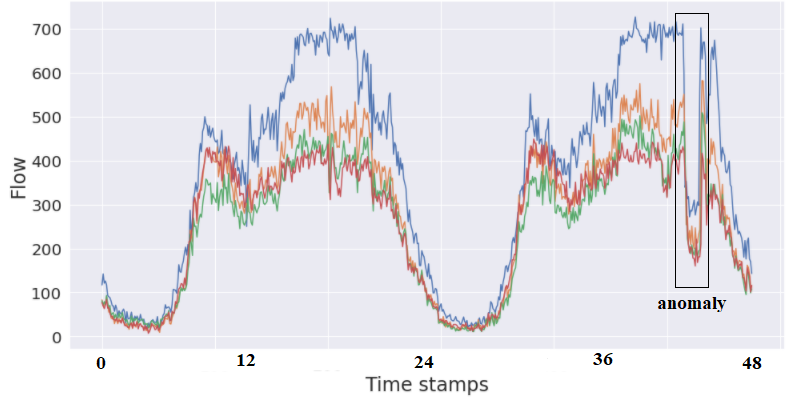}
  }
  \hspace{-0.0in}
  \caption{Representation of traffic flow for the first four selected sensors. The anomaly is represented with a box, as there is a big reduction in the flow.}\vspace{-0.1in}\label{fig::anomalies}
\end{figure}

The above representations shows spatio-temporal and spatial clusters. Our clusters are based on temporal similarity. If a data point is far from the mean of clusters, it means that the data point is rarely occur in temporal domain, or it is an anomaly. Here, we visualize such an example to clarify this interpretation. In Fig \ref{fig::mean_distance}, the heatmap of distance of data points from the center of clusters is represented and we can see that a portion of values are far form the centers. Time stamps close to 45 and the first 4 locations have light values. Regardless of the reason of anomaly, we look at traffic flow values in Fig. \ref{fig::anomalies}, for the first four location. The area close to time stamp 45 includes a big reduction in traffic flow values. This could be the result of an accident; however, in this paper, we do not analyze the reasons behind anomalies and the performance of anomaly detection. These analysis shows different application and importance of having spatial, temporal and spatio-temporal clusters in a transportation network.

\section{Conclusion and Future Work}

A spatio-temporal clustering is an important method for transportation systems. One of the challenging problems is to find spatio-temporal similarities in a transportation network. To obtain these similarities in traffic flow data, the problem definition is represented in Section II. Finding dynamic clusters of locations in a transportation network, illustrated in Section V.E, is necessary to analyze traffic congestion propagation and to improve traffic flow prediction and missing data imputation. Moreover, finding temporal patterns in traffic flow data is a method for more efficient prediction and detection of anomalies, illustrated in Section V.E. While these applications are important in transportation systems, there are few studies in the literature to develop deep learning models for spatio-temporal clustering of traffic flow data.

Increasing in the availability of traffic data requires further development of clustering models for complex and high-dimensional data. In this paper, We propose Spatial-DEC, a variation of Deep Embedded Clustering, to obtain spatio-temporal clusters, and illustrate its performance for finding dense and compact temporal clusters in Section V.C and spatially connected clusters in V.D. The contributions of this work are both in model architecture examined its validity in Section V.B and defining evaluation metrics for spatial and temporal clusters in Section V.C and V.D. The proposed model uses the loss function introduced in Eq. 6, and finds spatio-temporal clusters. Such a model can be useful not only for traffic data, but also for other spatio-temporal problems, such as environmental science and smart cities domains. Also, we consider a graph-structure for latent feature representation, which can be further studied in development of deep learning models for spatio-temporal data.

\bibliographystyle{ieeetr}%

\bibliography{sample-base.bib}

\end{document}